\RequirePackage{fix-cm}
\documentclass[twocolumn]{svjour3}          
\smartqed  
\usepackage{amsmath}
\usepackage{algorithm}
\usepackage{algorithmic}
\usepackage{graphicx}
\usepackage{color}
\usepackage{multirow}

\begin{document}

\title{Real-time and robust multiple-view gender classification using gait features in video surveillance
}



\authorrunning{Author} 

\institute{T.D. Do \at
              Hi-tech Bldg., Inha University, Inha-Ro 100, Nam-Gu, Incheon, South Korea. (402-751) \\
              Tel.: +82-32-860-7385\\
              Fax: +82-32-873-8970\\
              \email{dotd@inha.edu}           
             \emph{Present address:} of F. Author  
           \and
           V. H. Nguyen \at
           	  Faculty of Information Technology, Ton Duc Thang University, Ho Chi Minh City, Vietnam   \\           
              \email{nguyenvanhuan@tdt.edu.vn}
          \and
           H. Kim \at             
              Inha University, Inha-Ro 100, Nam-Gu Incheon, South Korea.	 \\
              \email{hikim@inha.ac.kr}
}


\maketitle

\begin{abstract}
It is common to view people in real applications walking in arbitrary directions, holding items, or wearing heavy coats. These factors are challenges in gait-based application methods because they significantly change a person’s appearance. This paper proposes a novel method for classifying human gender in real time using gait information. The use of an average gait image (AGI), rather than a gait energy image (GEI), allows this method to be computationally efficient and robust against view changes. A viewpoint (VP) model is created for automatically determining the viewing angle during the testing phase. A distance signal (DS) model is constructed to remove any areas with an attachment (carried items, worn coats) from a silhouette to reduce the interference in the resulting classification. Finally, the human gender is classified using multiple view-dependent classifiers trained using a support vector machine. Experiment results confirm that the proposed method achieves a high accuracy of 98.8\% on the CASIA Dataset B and outperforms the recent state-of-the-art methods.
\end{abstract}

\section{Introduction}
\label{intro}

Gender classification has an important role in modern society for surveillance or smart adaptation systems. It would be advantageous if a computer system or a machine could correctly classify an individual’s gender. For example, a surveillance camera system of mall shoppers could be beneficial to know the gender of the customers to create a proper strategy, or a sale-man robot \cite{ref0} could use an appropriate and smart approach to communicate with customers based on their gender.

Human gender, an active and promising area of research, can be classified using either a recorded voice \cite{ref1}\cite{ref2}\cite{ref3} or face image \cite{ref4}\cite{ref5}\cite{ref6}\cite{ref7}\cite{ref8}\cite{ref9}\cite{ref10}\cite{ref10a}. SexNet \cite{ref4} is an early system for gender classification using face images. The system uses the back-propagation algorithm of a neural network to train the gender classifier and obtains an error rate of 8.1\%. Based on this encouraging result, the system demonstrates that automatic gender classification by computers is feasible. However, the use of voice and face features for gender classification has limitations when the objects are distant from the sensor because it is difficult to obtain a high-quality recorded voice or face image from a distance.

Many psychological and medical experiments \cite{ref11}\cite{ref12} have indicated that humans and their gender can be recognized using their gait features. Therefore, gait features appear as alternative cues for resolving the recognition problem that occurs at long distances. Compared with other biometric features, gait information has particular advantages:
\begin{enumerate}
\item Easily obtainable from public areas and from a distance: Even when the subject is distant from the camera, we remain able to capture their gait information with an acceptable level of quality for specific tasks such as gait recognition and gender classification.
\item Uses simple instruments: Capturing the gait features requires only a simple conventional camera that can be placed anywhere in public areas such as banks, parking lots, and airports.
\item Does not require collaboration with the subjects: Gait features can be captured easily, even without the subject’s permission. Although this is an advantage, it raises the issue of the right to privacy.
\item It is difficult to forge or falsify gait features: Gait features indicate the walking manner of a human, characterizing their physical capability. Mimicking the gait of other people is difficult.
\end{enumerate}

\begin{figure}[!b]
  \centering
  \includegraphics[scale=0.35]{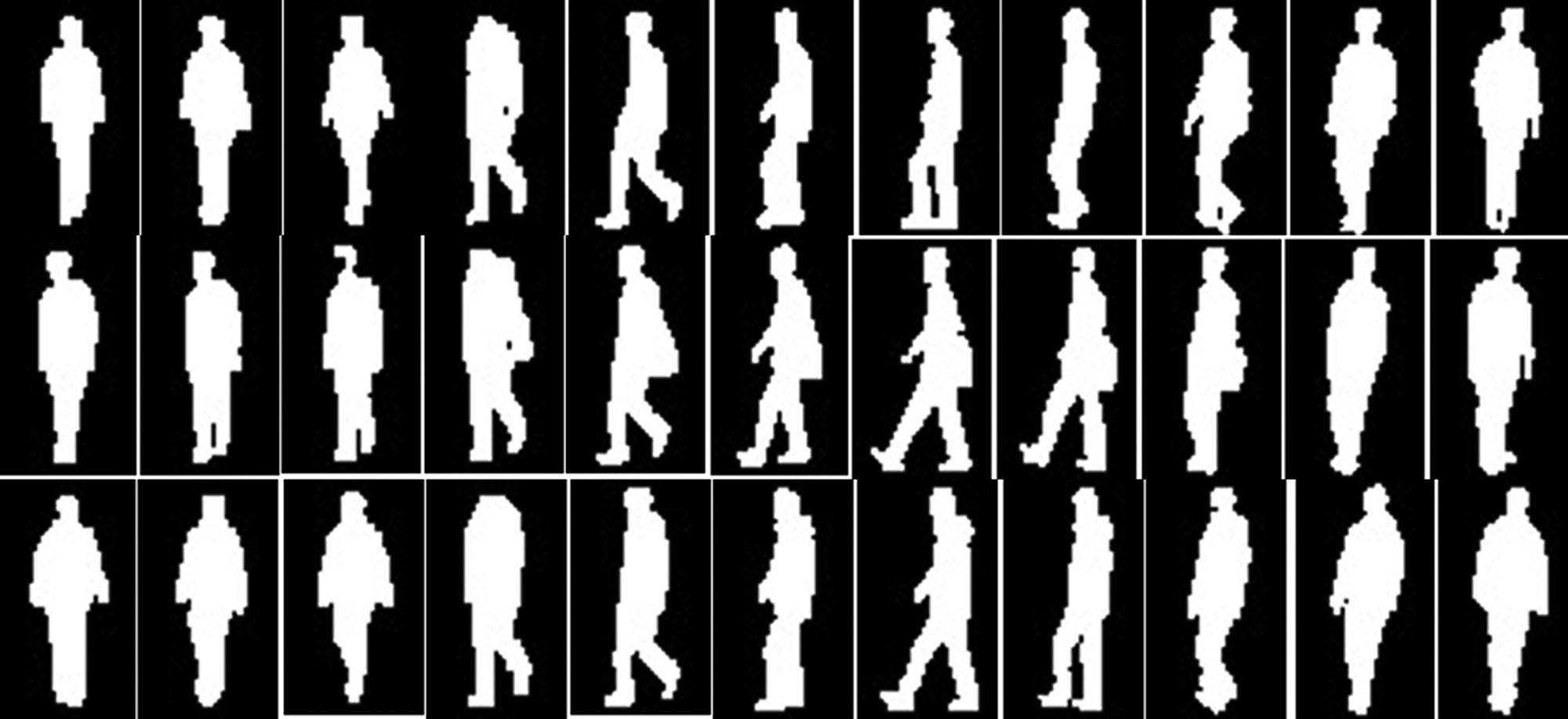}
\caption{Challenges in gait analysis of humans: change in viewpoint (top row), carrying an item (middle row), wearing a coat (bottom row)}
\label{fig:1}       
\end{figure}

However, gait-based systems such as gender classification and gait recognition share the same challenges as indicated in Fig.~\ref{fig:1}. These challenges arise from the environment including the viewpoint of camera changes or the subject’s physical characteristics such as carrying a backpack, wearing a heavy coat, or displaying signs of an injury. These factors change the subject’s appearance leading to a significant effect on their gait information as they move \cite{ref13}\cite{ref14}.


\begin{figure}[!b]
\centering
\includegraphics[scale=0.25]{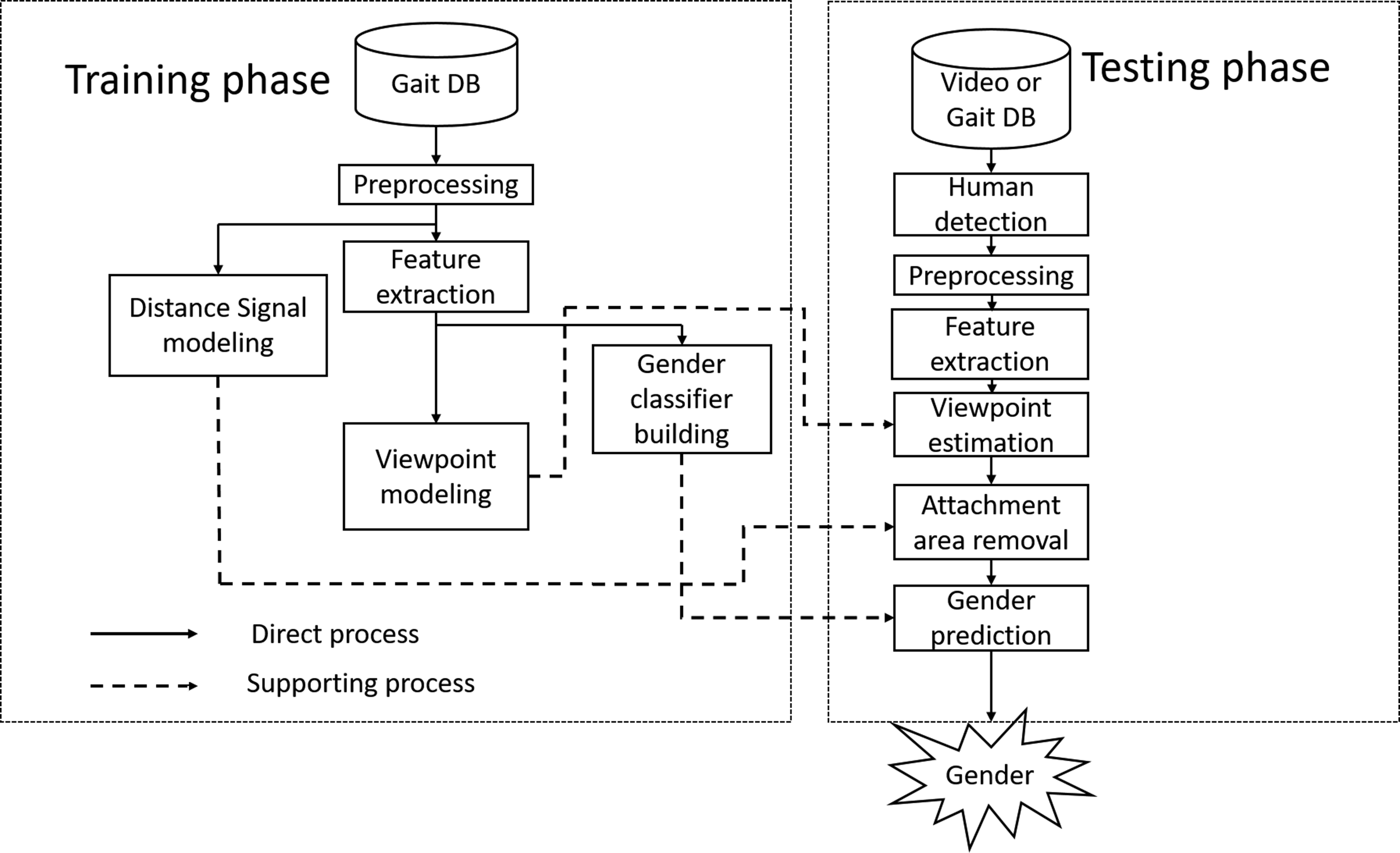}
\caption{Flowchart of proposed gender-classification process}
\label{fig:2}
\end{figure}

This paper proposes a novel gender classification method for use with an arbitrary viewpoint. To improve the performance, we present a method to remove areas with an attachment, such as a heavy coat or backpack. A general flowchart of the proposed method is presented in Fig. ~\ref{fig:2}, which includes two major phases: training and testing. During the training phase, after the preprocessing step, the distance signal (DS) model, viewpoint (VP) model, and view-dependent gender classifier are built. Before building the VP model, the average gait image (AGI) and the lower portion of the average gait image (LAGI) are generated. During the testing phase, after the human detection and preprocessing step, the viewpoint of the current object is estimated using the current silhouette image and the VP model from the training phase. The attachment-area removal module is then used to eliminate unwanted areas such as backpacks or bags to obtain an attachment-free silhouette. Based on the estimated viewpoint of the current object, the corresponding classifier of that viewpoint (built during the training phase) is applied to the attachment-free silhouette to classify the object gender. The contributions of this paper are as follows:
\begin{itemize}
\item Building a VP model for viewpoint estimation, allowing the proposed method to estimate the viewing direction automatically.
\item Building a DS model for attachment-area removal to eliminate the noise generated from carried objects, which significantly degrades the performance of the system.
\item Building a viewpoint-dependent gender classifier using an SVM \cite{ref15} that allows the algorithm to function from any viewpoint.
\end{itemize}

This paper is organized as follows. Section 2 discusses related works. Section 3 introduces the proposed method including the training and testing phases. The details of the key modules, such as VP modeling and estimation, DS modeling, viewpoint-dependent classifier building, and attachment-area removal are discussed in this section. A pseudo-code of the overall flowchart is also presented in this section. Section 4 presents the experimental results on public datasets. Finally, Section 5 concludes this paper and provides some areas for future work.

\section{Related works}
Gait-based recognition techniques can be divided into two categories, marker-based and markerless methods. Using markers, in an early work, Kozlowski and Cutting \cite{ref16}\cite{ref17} attempted to attach a point-light display (marker-based method) to a human body to extract the gait information. With this system, a human observer can determine a subject’s gender based on the signals obtained with an acceptable level of accuracy (63\%). However, to capture the gait information, the subject is required to wear a swimsuit and special devices, which is inconvenient, unfriendly, and impractical in real circumstances.

Today, owing to technical innovations in camera and sensor development, the human gait can be easily obtained without a point-light display, leading to the development of markerless methods. The markerless-based methods for gait recognition can be classified into the model and appearance-based approaches. Such categorization can also be used for gender classification.

In the model-based methods \cite{ref18}\cite{ref19}\cite{ref20}, the human body is divided into various parts, the structures of which are then fitted using primitive shapes such as ellipses, rectangles, and cylinders. Then, the gait feature is encoded using the parameters of the primitive shapes to measure the time-varying motion of the subject. In \cite{ref18}, L. Lee et al. divide a human silhouette into seven different parts corresponding to the head and shoulder region, the front of the torso, back of the torso, front thigh, back thigh, front calf and foot, and back calf and foot. They then use ellipses to fit the model and capture the parameters of the ellipses such as the mean, standard deviation, orientation, and magnitude of the major components as feature vectors for classification. Although such methods are robust to noise and occlusions, they typically require a relatively high computational cost.

Appearance-based methods \cite{ref21}\cite{ref22}\cite{ref23}\cite{ref24}\cite{ref25}\cite{ref26} analyze the spatio-temporal shape and dynamic motion characteristics of the silhouette in a gait sequence without using a human body model. A gait energy image (GEI) \cite{ref21} is frequently used to encode the gait features because it includes both static (body shape) and dynamic information (arm swings and leg movements). The GEI feature is defined as the average frame of the subject in the gait cycle. Compared to the model-based methods, the appearance-based methods are considerably faster. In \cite{ref23}, instead of modeling the silhouette, Shiqi Yu et al. calculate the GEI and use it to create the seven-part model defined in \cite{ref17}. Because the contribution of each part to the gender classification varies, the authors assign different weights to the parts based on their experiments. Such methods obtain highly accurate classification rates (approximately 95\%). However, they were developed to the only function on a side view, making it inappropriate to apply in real applications.

In \cite{ref13}\cite{ref14}\cite{ref27}\cite{ref28}\cite{ref29}\cite{ref30}\cite{ref31}\cite{ref32}\cite{ref47}, the researchers attempted to resolve gender classification from multiple viewpoints. In \cite{ref31}\cite{ref32}, De Zhang et al. build an invariant classifier by combining the GEIs of different viewpoints into a single third-order tensor. They then use multiple linear principal component analysis (PCA) to reduce the dimensions and apply a support vector machine (SVM) to create a discriminative gender classifier. From another perspective, Kale et al. \cite{ref33} use complicated equations from the structure of motion \cite{ref34} to eliminate the viewpoint effect by synthesizing the side view from other viewpoints. A final recognition task is conducted on the synthesized data. Issac et al. \cite{ref47} propose a method to delineate the gait instance as a sequence of poses or frames based on the fact that humans tend to assume certain poses at each part of a gait cycle. The gender of each frame is predicted, and the gender decision of a sequence is then made using majority voting. However, none of the previous works considers solving the problem of a subject carrying an item or wearing a heavy coat, which are common situations in real applications that can significantly degrade the classification rate.

\section{Proposed method}
\label{sec:3}

\subsection{Dataset and preprocessing step}
\label{sec:31}
This paper proposes a method for gender classification from an arbitrary viewpoint, and therefore, a dataset with multiple camera views is required. For this purpose, the CASIA gait Dataset B \cite{ref13}\cite{ref14} was utilized throughout this study for illustrative and experimental purposes.

A person is first detected using a histogram of oriented gradient (HOG) \cite{ref35}. During the preprocessing step, a classic background subtraction \cite{ref36} is then applied to obtain a person’s silhouette. Because a person’s size changes from frame to frame, it is necessary to normalize the human bounding box before the training process. Assume in frame $I_t$ at time $t$, that a human is detected with a bounding box $B_t$; denote their silhouette obtained from the background subtraction as $S_t$. To register the silhouette image, the center of the silhouette $P_t$ (reference point) at time t is computed as:
\begin{equation}
\label{eq1}
P_t(x_0, y_0) = \left(\frac{M_{10}}{M_{00}}, \frac{M_{01}}{M_{00}}\right)
\end{equation}
where $M_{ij}$ are the raw moments of the binary silhouette image defined by $M_{ij}=\sum_{x}\sum_{y}x^iy^jS_{t}(x,y)$.

The silhouette image $S$ is resized to the fixed height $h$ to maintain the human ratio scale. The resized image is then zero padded or cropped on both sides (left and right sides) to ensure that the silhouette image has the predefined width $w$.

\begin{figure*}
\centering
\includegraphics[width=1\textwidth]{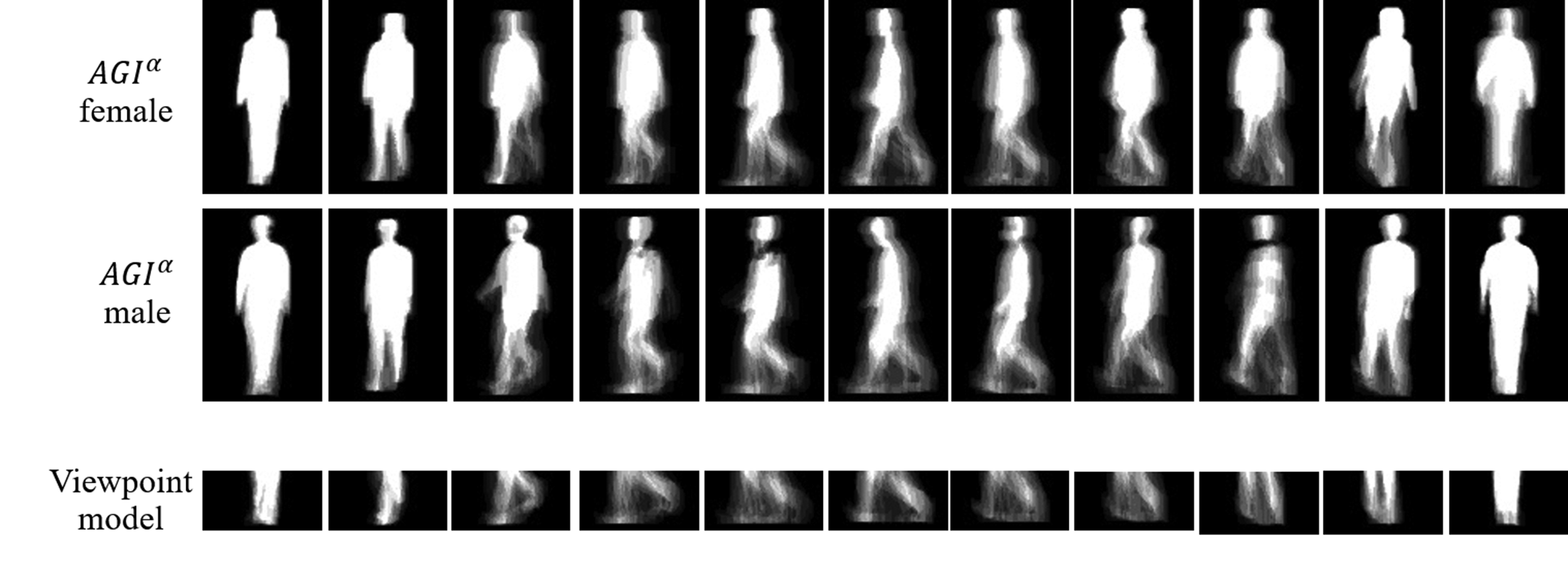}
\caption{Example AGIs: models from different viewpoints}
\label{fig:3}
\end{figure*}

\subsection{Viewpoint modeling and estimation}
\label{sec:32}
\subsubsection{Viewpoint modeling}
\label{sec: 321}
A person can move in an arbitrary direction under real circumstances. Therefore, in this paper, rather than using GEI \cite{ref21} as a representation of a gait feature for classification, AGI is defined, as shown in Fig.~\ref{fig:3}. The main difference between GEI and AGI is that the gait cycle information, which must be calculated into a GEI, is not required in an AGI. Furthermore, applying the gait cycle during the feature extraction step makes the entire algorithm inflexible because the gait cycle can be calculated accurately only when the person is captured from a side view, which is impractical under real circumstances. With the proposed method, all models are trained based on each viewpoint independently. Subsequently, we describe the details required for training for viewpoint $\alpha$ using the data on that viewpoint. The AGI is defined as:
\begin{equation}
\label{eq2}
AGI^{\alpha}(x, y)=\frac{1}{T}\sum_{t=1}^{T}S_t^\alpha(x,y)
\end{equation}
where $T$, gait period, is defined adaptively using the video frame rate $f$ and approximate gait cycle time $\mu$ as $T = \mu*f$. According to \cite{ref37}\cite{ref38}, when the frame rate $f$ is 25 frames/s, the value of the gait cycle time $\mu$ must be 0.6 seconds to capture the most informative gait features; thus, $T = 0.6*f$ is used in the proposed method; $S_t^\alpha$ is the silhouette image at time $t$ with the viewpoint $\alpha$.

Defining $\gamma_k^\alpha$ as $\gamma_k^\alpha=\left\{{AGI}_1^\alpha, {AGI}_2^\alpha, ..., {AGI}_n^\alpha\right\}_k$, is the feature vector of subject $k$ in a viewpoint $\alpha$, $\alpha=\overline{1, \nu}$ and $k=\overline{1,N}$, where $\nu$ and $N$ are number of viewpoints and number of subjects (training samples), respectively. In fact, for the training step, a greater number of training samples $N$ is preferable. The corresponding label of $\gamma_k^\alpha$, denoted as set $L_k^\alpha=\{y_1^\alpha, y_2^\alpha, ..., y_n^\alpha\}_k, k=\overline{1, N}; \alpha=\overline{1,\nu}; y_i^\alpha\in\{-1,1\}$, is used to indicate the gender ("-1" for female and "1" for male).

To estimate the viewpoint for the input during the testing phase, we construct a viewpoint model $D$. This viewpoint model includes the viewpoint templates of an individual view. The viewpoint template is calculated as the average silhouette of all sequences from the $\alpha$-th viewpoint. As observed, the viewpoint is clearly distinguished in the lower part of the silhouette and therefore, the $\alpha$-th viewpoint template, denoted as ${LAGI}^\alpha$, is extracted as the lower part of the average silhouette denoted as $LPS^\alpha$ with a height of 0.715$h$ to $h$, as suggested by \cite{ref39}, where $h$ is the height of the silhouette:

\begin{equation}
\label{eq3}
{LPS}^\alpha(x,y)=S^\alpha(x,y), x=\overline{0.715h, h}, y=\overline{1, w}
\end{equation}
\begin{equation}
\label{eq4}
{LAGI}^\alpha(x,y) = \frac{1}{N}\sum_{t=1}^N{LPS_t^\alpha(x,y)}
\end{equation}

The viewpoint model is then denoted as $D_{Low}=D=\{{LAGI}^0,{LAGI}^1, …,{LAGI}^\nu \}$, where $\nu$ is the number of viewpoints. This viewpoint model $D_{Low}$ is used to estimate the viewpoint of a person walking during the testing phase.

\subsubsection{Viewpoint estimation}
With this method, the attachment-area removal module and gender classifier are dependent on the viewpoint; thus, the viewpoint is first estimated. During the testing phase, given the sequences of the silhouettes, the average gait image of the current walking subject, ${AGI}^c$, is calculated using Eq.~\ref{eq2}. Then, ${LAGI}^c$  is extracted from the lower part of ${AGI}^c$  based on the size given in Eq.~\ref{eq3}, rather than recalculating $LAGI$, as during the training phase.

\begin{figure*}
\centering
\includegraphics[width=1\textwidth]{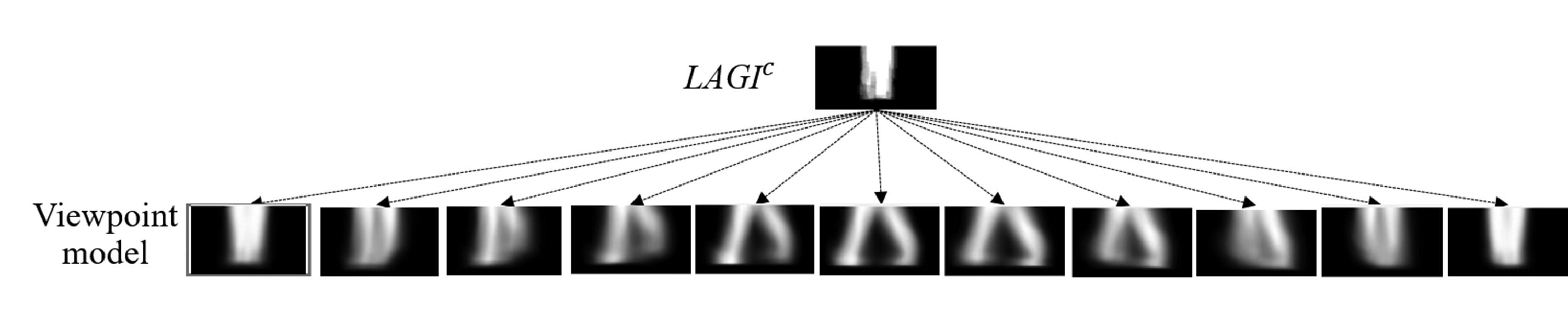}
\caption{Viewpoint model of 11 viewpoint templates where ${LAGI}^c$ is matched with the first template}
\label{fig:4}
\end{figure*}

To obtain viewpoint $\alpha$ of the current walking subject, ${LAGI}^c$  is matched with each viewpoint template in viewpoint model $D_{Low}$ using the Euclidean distance. The least distance is then selected for the viewpoint estimation, as indicated in Fig. ~\ref{fig:4}.
\begin{equation}
\label{eq5}
\alpha=\min_j||{LAGI}^c-{LAGI}^j||_2, j=\overline{0,\nu}
\end{equation}

\subsection{Distance signal modeling and attachment removal}
\subsubsection{Distance modeling}
In real applications, it is common to view people moving with attached objects such as bags or backpacks; similarly, their appearance can be significantly changed when wearing a heavy coat. The added area resulting from a held item or worn coat contributes nothing to the result of the gender classification. In actuality, these factors negatively influence the results of the classification. In this section, a distance signal (DS) model of humans under normal walking conditions (not holding anything and wearing thin clothes) from different viewpoints is proposed for removing these redundant attachments.

Given a set of silhouettes in movement direction $\alpha$ , for each silhouette, a distance signal is built. Considering the mass reference point calculated by Eq. ~\ref{eq1}, each point $P_i$  on the silhouette boundary is represented in polar coordinates by two parameters, $d_i$ and $\theta_i$, which indicate the distance from the point $P_i$ to the reference point $P$, and the angle formed by the line connecting the point to the reference point ${PP}_i$ with the horizon ${PP}_h$, respectively. 

\begin{figure*}
\centering
\includegraphics[width=1\textwidth]{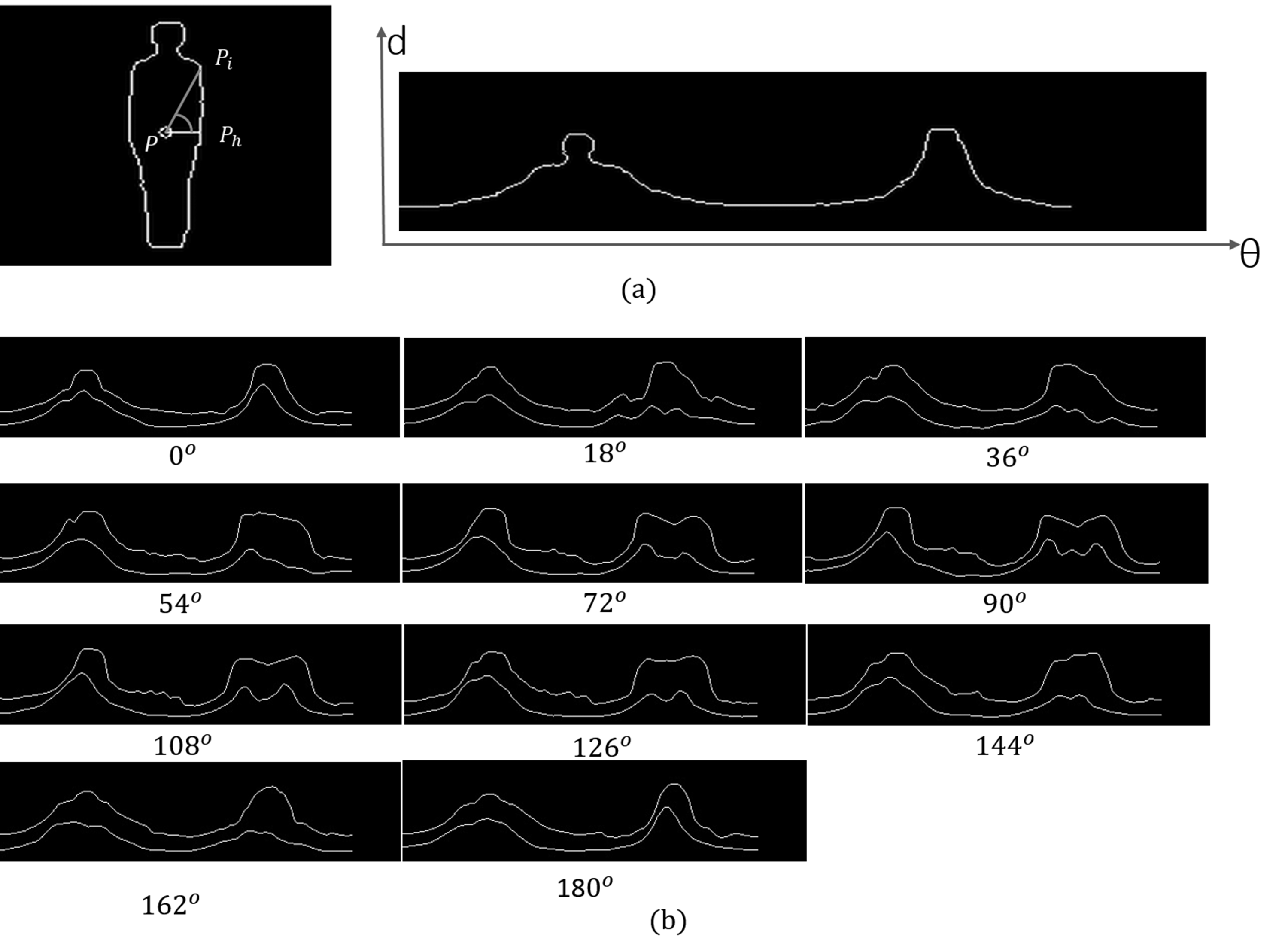}
\caption{Example of distance signal (a) and distance signal model for different viewpoints (b)}
\label{fig:5}
\end{figure*}

\begin{equation}
\label{eq6}
d_i=||P-P_i||_2
\end{equation}

\begin{equation}
\label{eq7}
\theta_i=\arccos\left(\frac{\overrightarrow{PP_i}*\overrightarrow{PP_h}}{|\overrightarrow{PP_i}||\overrightarrow{PP_h}|}\right)
\end{equation}
where * is the dot product between two vectors and the value of angle $\theta$ varies from $0^o$ to $360^o$ computed counterclockwise; $P_i$ and $P_h$ are depicted in Fig. \ref{fig:5}a (left). The DS signal is then constructed by continuously concatenating these parameters from $P_h$ counterclockwise to define the signal presented in Fig. \ref{fig:5}a (right). After building these DS signals for viewpoint $\alpha$, denoted by ${DS}^\alpha=\{{DS}_1^\alpha,…,{DS}_n^\alpha\}$, the DS model for viewpoint $\alpha$ is constructed using two curves, $MaDS^\alpha$ and $MiDS^\alpha$, which are defined as:

\begin{equation}
\label{eq8}
MaDS^\alpha=\{d_i^{max}, \theta_i\}, \text{where } d_i^{max} = \max\limits_{k=\overline{1,N}}\{d_k|\theta_k\}
\end{equation}

\begin{equation}
\label{eq9}
MiDS^\alpha=\{d_i^{min}, \theta_i\}, \text{where } d_i^{min}=\min\limits_{k=\overline{1,N}}\{d_k|\theta_k\}
\end{equation}

The distance signals are smoothed using the moving average technique with the number $n_{avg}=3$ before calculating the DS model. Fig. \ref{fig:5}b illustrates the DS model for 11 viewpoints in our experiments.

\subsubsection{Attachment-area removal}
During the testing phase, the DS signal of the silhouette of the current subject $DS^c$ is calculated in the manner described in the section 3.3.1. Given the viewpoint estimated using the viewpoint estimation module, the current $DS^c$ is projected to the corresponding viewpoint DS model. The current $DS^c$  is modified using the following rule to eliminate any attachments, if they exist:

\begin{equation}
\label{eq10}
DS^c=\{d_i^c, \theta_i\} \text{where } d_i^c= \begin{cases} d_i^c & \text{if }  d_i^c \leq d_i^{max} \\
d_i^{min} & \text{ otherwise}
\end{cases} 
\end{equation}

\begin{figure*}
\centering
\includegraphics[width=1\textwidth]{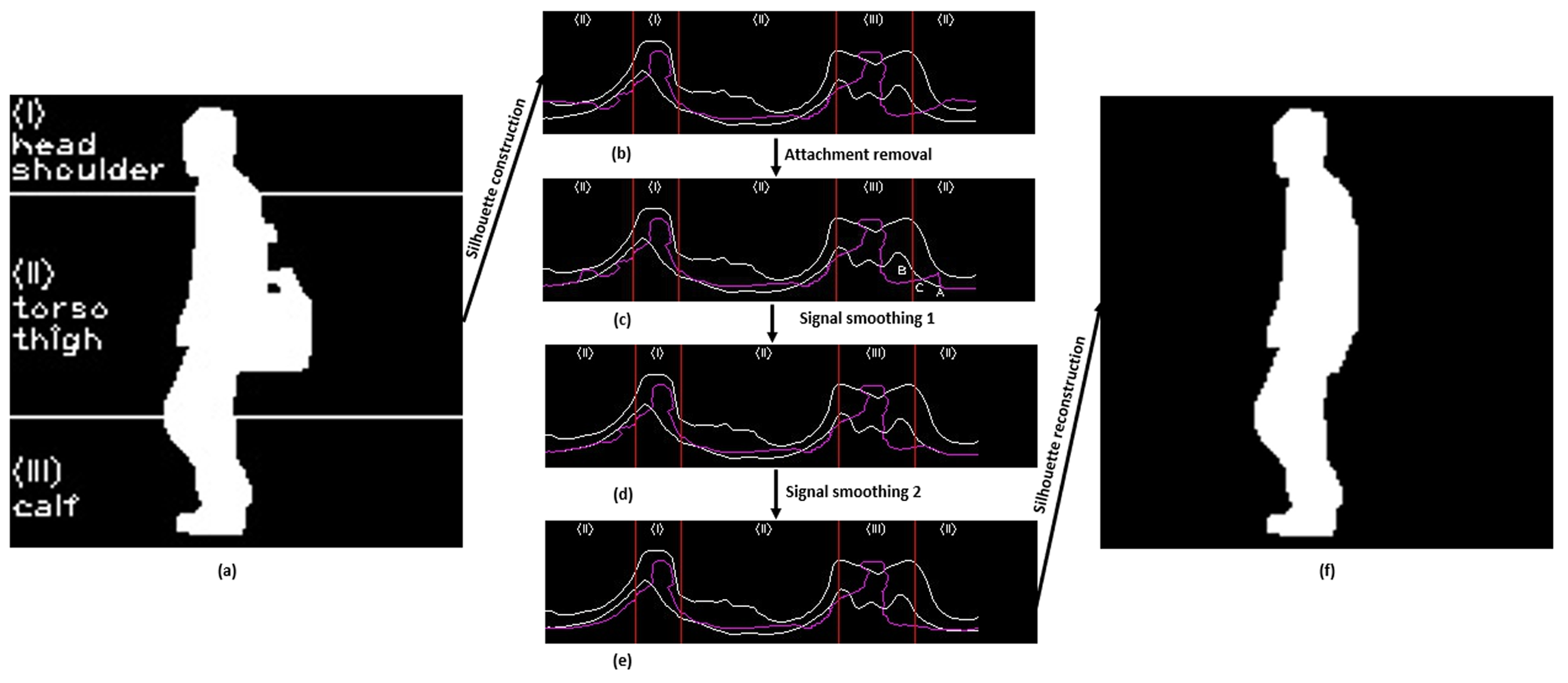}
\caption{Example of original silhouette, DS of the silhouette, corrected DS of the silhouette, and silhouette reconstruction of the current subject from a side view}
\label{fig:6}
\end{figure*}

Fig. \ref{fig:6} illustrates this process. In the figure, an example of a side-view silhouette input with an attachment is provided because in this view the attachment can be seen most clearly. Our goal is to remove the attachment from the human silhouette, therefore, the human silhouette is divided into three parts, as shown in Fig. \ref{fig:6}a. The first part consists of head and shoulder with a height of 0 to 0.17$h$, as suggested in \cite{ref45}\cite{ref46}, where $h$ is the height of human silhouette. The second part includes the human torso and thigh with a height of 0.17$h$ to 0.715$h$. Finally, the last part is human calf with a height of 0.715$h$ to $h$, as suggested in \cite{ref39}. The red vertical lines (Fig. 6b, c, d, e ) are drawn to separate these three parts from the human silhouette. Firstly, the input silhouette is converted into a distance signal $DS^c$, presented as violet curve in Fig. \ref{fig:6}b. The white curves are the maximum distance signal $MaDS^\alpha$ (upper) and minimum distance signal $MiDS^\alpha$ (lower) from the DS model for a specific viewpoint ($\alpha=90^o$).

When human carrying a backpack or bag, the appearance of torso thigh part is
changed due to the attachment, thus only this part is taken into account for
correction. As we observed from the experiments, a $DS^c$ with a value less than
$MiDS^\alpha$ is the noise from an imperfect background subtraction. A $DS^c$
with a value greater than $MaDS^\alpha$ is considered as the attachment area
from a subject carrying an item while walking. Using the DS model from
Section 3.3.1, the attachment area can therefore be removed. The corrected
version of $DS^c$ is obtained by replacing the violated signal with the 
corresponding values of the $MiDS^\alpha$ curve at Point A, as shown in Fig. 
\ref{fig:6}c. To avoid the problem of strict change in the resulted signal, we 
continue to look from the point A to point B in Fig. \ref{fig:6}c to find the 
point that has the smallest vertical distance between $MiDS^\alpha$ and the 
resulted signal (point C). The segment of the resulted signal from A to C is again 
replaced by $MiDS^\alpha$, as presented in Fig. \ref{fig:6}d. The signal in Fig. 
\ref{fig:6}e is obtained by smoothing using the average filter $\frac{1}{5}[1 1 1 1 1]$. The same process is applied to all segments of torso/thigh part of the $DS^c$ curve. Finally, the corrected version of the $DS^c$ is used to reconstruct the silhouette of attachment-free, as shown in Fig. \ref{fig:6}f. The updated version of the silhouette is then used to recalculate the AGI for gender classification.

\subsection{Gender classifier building}
SVM \cite{ref15} is a superior tool for a binary classification problem regarding minimizing the classification error and maximizing the margin between the two classes. Because gender classification is a binary classification task, a standard SVM with a linear kernel was selected to train the view-dependent classifiers. For solving a constrained quadratic optimization problem, we set the maximum number of iteration to 100.

To create the viewpoint-dependent classifier, the feature sets $\gamma^\alpha=\{\gamma_1^\alpha,\gamma_2^\alpha, …,\gamma_N^\alpha\}$ and its corresponding labels $L^\alpha=\{L_1^\alpha,L_2^\alpha, …,L_N^\alpha\}$ are used as inputs for the linear SVM. The $\alpha$-th viewpoint classifier, obtained by using the SVM, is denoted as  $C_{gen}^\alpha$. The multiple-view classifier is a collection of different viewpoint-dependent classifiers, which is denoted as $C_{gen}={C_{gen}^\alpha}$ and $\alpha=\overline{1, \nu}$. 

In the testing phase, the viewpoint is estimated as discussed in section 3.2.2. Based on the estimated viewpoint $\alpha$, the corresponding classifier $C_{gen}^\alpha$ is automatically selected from  $C_{gen}$ to predict the person’s gender in a current frame.

The algorithms 1 and 2 give a more detail description of the proposed method using pseudo-code in which all notations described above are used.

\begin{algorithm*}
\caption{The training phase}
\begin{algorithmic}[1]
\STATE\textbf{Input}: $video\texttt{\_}sequences$ in the $training\texttt{\_}samples$ of male and female in all viewpoints 
\STATE \textbf{Output}: VP model, DS model, and $C_{gen}$ classifier

\FORALL {$\alpha$ \textbf{in} $views$}
	\FORALL {$video\texttt{\_}sequence\; k$ \textbf{in} $training\texttt{\_}samples$ of viewpoint $\alpha$ }
		\FORALL {frame $t$ \textbf{in} $video\texttt{\_}sequence$}
			\STATE Human detection
			\IF {no human detected}
				\STATE Skip to the next frame
			\ENDIF
			\STATE Preprocessing to get the normalized silhouette $S^\alpha$
			\STATE Extract low part of the silhouette $S^\alpha$ by Eq.~\ref{eq3} and accumulate to $LPS_{k,t}^\alpha$
			\STATE Detect contour of a normalized silhouette $S^\alpha$
			\STATE Calculate  $d_i$ and $\theta_i$ by Eq.~\ref{eq6}, Eq.~\ref{eq7}
			\STATE Calculate $AGI^\alpha$ and assign its label $y^\alpha$ by Eq.~\ref{eq2}
			\STATE Append $AGI^\alpha$ to vector $\gamma_k^\alpha$
			\STATE Append $y^\alpha$ to vector $L_k^\alpha$
		\ENDFOR
	\ENDFOR
	\STATE Calculate the $LAGI^\alpha$ using $LPS_{t,f}^\alpha$ by Eq.~\ref{eq4} for VP model
	\STATE Calculate $MaDS^\alpha$ and $MiDS^\alpha$ using $d_i$ and $\theta_i$ by Eq.~\ref{eq8}, Eq.~\ref{eq9} for DS model
	\STATE Train view-dependent classifier $C_{gen}^\alpha$ using $\gamma^\alpha$ and $L^\alpha$ as inputs of SVM
\ENDFOR
\end{algorithmic}
\end{algorithm*}

\begin{algorithm*}
\caption{The testing phase}
\begin{algorithmic}[1]
\STATE \textbf{Input}: $video\texttt{\_}sequence$ of a person in an unknown viewing angle
\STATE \textbf{Output}: Gender information
\STATE Initialize $counter = 0$
\STATE Initialize empty vector $v$
\FORALL {frame $t$ \textbf{in} $video\texttt{\_}sequence$}
	\STATE Human detection
	\IF {human detected}
		\STATE Increase $counter$ by 1
	\ELSE
		\STATE $counter=0$
		\STATE Skip to the next frame
	\ENDIF
	\STATE Preprocessing to get the normalized silhouette $S$
	\STATE Append $S$ to the end of vector $v$
	\IF {$counter\geq15$}
		\STATE Calculate the $AGI$ using a vector of silhouette $v$
		\STATE Extract low part of average gait image of the current frame $LAGI^c$ from $AGI$
		\STATE Estimate the viewpoint $\alpha$ using $LAGI^c$ and VP model by Eq.~\ref{eq5}
		\STATE Remove the attachment area using estimated viewpoint $\alpha$ and DS model by Eq.~\ref{eq10}
		\STATE Reconstruct silhouette $S\rightarrow S$' and update $AGI\rightarrow AGI$'
		\STATE Predict the gender in current frame using the updated $AGI$' and the estimated viewpoint $\alpha$
		\STATE Remove the first element of $v$
	\ENDIF
\ENDFOR
\end{algorithmic}
\end{algorithm*}

\section{Experimental results}
\subsection{Experimental dataset}
The CASIA Dataset B \cite{ref13}\cite{ref14} addresses our requirements of multiple camera views because it includes sequences of various people from 11 viewpoints (from $0^o$ to $180^o$) under different walking conditions such as walking normally, carrying a backpack, and wearing a coat. The CASIA Dataset B captures sequences of 124 individual people (31 females and 93 males). Each person is captured ten times to create ten different sequences including six sequences under normal walking conditions, two backpack-carrying sequences, and two coat-wearing sequences. Table~\ref{tab:1} summarizes the information of the CASIA Dataset B.


%
%

\begin{table}[!b]
\centering
\caption{Summary of CASIA Dataset B}
\label{tab:1}       
\begin{tabular}{lll}
\hline\noalign{\smallskip}
\text{Walking condition} & \text{\#subjects} & \text{\#sequences}  \\
\noalign{\smallskip}\hline\noalign{\smallskip}
\text{Normal walking} & 6 & $6\times124\times11$ \\
\text{Carrying a bag} & 2 & $2\times124\times11$ \\
\text{Wearing a coat} & 2 & $2\times124\times11$ \\
\noalign{\smallskip}\hline
\end{tabular}
\end{table}

The CASIA Dataset B includes background subtraction and thus, in the proposed system we are only required to resize and center the silhouette to the same size (144$\times$144). For the AGI calculation, we must accumulate fifteen frames; it requires approximately 0.6 seconds to obtain the first gender-classification result when the frame rate is 25 fps, which can be considered a system delay.

We used the CASIA Dataset B for both training and testing using the same protocol as in \cite{ref32}, which uses n-fold cross-validation. With this protocol, all 31 females were selected; 31 males were selected randomly from the CASIA Dataset B owing to a bias in the number of males in the dataset. The 31 females and 31 males were then grouped into 31 disjoint sets consisting of one female and one male. To create viewpoint-dependent classifiers, we use 30 sets for training. The remaining sets were used to test the system accuracy. The training and testing phases were repeated 31 times; the averages of the correct classification rate are listed for all experiments.
\subsection{Viewpoint-dependent classifiers test}
This test was used to validate the performance of only viewpoint-dependent classifiers under the assumption that the viewpoint was given. We conducted the test for both correct and incorrect viewpoint classifiers with respect to a specific viewpoint to observe the effect of viewpoint changes on the gender classification. Table \ref{tab:2} displays the correct classification rates ($CCRs$) when using the corresponding classifier and a non-corresponding classifier (the viewpoint is given). The $CCR$ is defined as:
\begin{equation}
\label{eq11}
CCR=\frac{TP+TN}{N}
\end{equation}
where $TP$ is the true positive referring to the cases in which the system correctly classifies positive samples (male to male), $TN$ is the true negative referring to the cases in which the system correctly classifies negative samples (female to female), and $N$ is the total number of samples. In these experiments, the male samples are labeled as 1 (positive samples) and the female samples are labeled as -1 (negative samples). As indicated in Table \ref{tab:2}, applying a proper classifier for a specific viewpoint provides higher $CCRs$ (97.6\% $\pm$ 0.881) (for further description see Table \ref{tab:2}).

\begin{table*}
\centering
\caption{CCRs (\%) of viewpoint-dependent classifiers for specific viewing angle under normal walking conditions}
\label{tab:2}       
\begin{tabular}{llllllllllll}
\hline\noalign{\smallskip}
\multirow{2}{*}{\text{Ground-truth viewpoint}} & \multicolumn{11}{c}{Classifier} \\
& $0^o$ & $18^o$ & $36^o$ & $54^o$ & $72^o$ & $90^o$ & $108^o$ & $126^o$ & $144^o$ & $162^o$ & $180^o$  \\
\noalign{\smallskip}\hline\noalign{\smallskip}
$0^o$ &97.6&96.5&94.1&89.6&89.9&82.3&87.2&93.8&94.6&94.5&95.8 \\
$18^o$&97.5&98.6&96.8&93.4&89.1&85.4&88.6&92.4&93.7&95.2&96.4  \\
$36^o$&95.3&96.6&97.4&95.8&94.2&93.4&93.9&94.5&95.7&95.7&94.2  \\
$54^o$&92.2&95.3&96.1&96.6&95.8&94.6&93.4&95.0&95.5&94.6&93.1 \\
$72^o$&92.3&93.2&94.1&95.4&96.1&94.8&94.5&95.2&93.0&92.8&92.4 \\
$90^o$&90.0&93.1&95.3&95.5&95.7&98.8&96.9&95.7&94.1&92.5&91.1 \\
$108^o$&91.4&92.7&95.4&96.1&96.5&97.0&97.3&96.2&95.5&92.4&93.2 \\
$126^o$&92.5&94.3&96.7&95.7&95.3&94.1&93.4&96.8&94.5&94.7&94.5 \\
$144^o$&95.7&96.8&97.8&95.8&93.2&93.4&93.9&94.8&97.5&95.5&94.5 \\
$162^o$&95.5&96.3&95.8&93.4&91.3&90.4&92.6&94.4&95.4&98.3&96.6 \\
$180^o$&96.8&97.6&95.8&92.8&91.5&91.4&92.6&92.3&93.5&95.3&98.5 \\
\noalign{\smallskip}\hline
\end{tabular}
\end{table*}

\begin{table*}
\centering
\caption{CCRs (\%) of viewpoint-dependent classifiers of specific viewing angle when carrying a bag}
\label{tab:3}       
\begin{tabular}{llllllllllll}
\hline\noalign{\smallskip}
\multirow{2}{*}{\text{Ground-truth viewpoint}} & \multicolumn{11}{c}{Classifier} \\
 & $0^o$ & $18^o$ & $36^o$ & $54^o$ & $72^o$ & $90^o$ & $108^o$ & $126^o$ & $144^o$ & $162^o$ & $180^o$  \\
\noalign{\smallskip}\hline\noalign{\smallskip}
$0^o$&94.3&91.6&89.7&84.8&81.5&79.6&80.3&83.9&85.2&89.9&92.6 \\
$18^o$&92.8&94.4&93.5&89.9&83.1&80.1&82.8&86.4&85.4&92.2&91.7  \\
$36^o$&89.8&92.1&93.7&90.7&85.2&82.0&88.8&90.2&86.4&90.5&89.1  \\
$54^o$&88.0&88.3&90.1&91.2&89.1&84.5&89.1&90.6&87.9&88.3&86.4 \\
$72^o$&88.1&89.4&89.5&90.4&90.6&86.2&89.4&89.1&89.3&88.8&84.4 \\
$90^o$&82.3&82.8&83.1&85.5&86.4&87.4&87.1&85.3&85.1&83.6&82.1 \\
$108^o$&82.2&82.4&83.9&84.2&85.5&86.7&89.8&87.4&87.0&85.4&83.7 \\
$126^o$&87.4&87.3&88.3&90.7&83.3&85.5&88.1&91.2&90.5&89.7&88.5 \\
$144^o$&88.7&89.3&90.6&90.1&82.2&83.4&87.3&90.2&91.4&89.8&88.4 \\
$162^o$&91.4&91.2&90.8&89.4&81.1&82.1&85.6&89.6&90.7&93.1&92.2 \\
$180^o$&93.2&92.5&91.0&90.7&80.5&80.6&84.1&87.1&89.1&91.4&94.8 \\
\noalign{\smallskip}\hline
\end{tabular}
\end{table*}

\begin{table*}
\centering
\caption{CCRs (\%) of viewpoint-dependent classifiers of specific viewing angle when wearing a coat }
\label{tab:4}       
\begin{tabular}{llllllllllll}
\hline\noalign{\smallskip}
\multirow{2}{*}{\text{Ground-truth viewpoint}}& \multicolumn{11}{c}{Classifier} \\
& $0^o$ & $18^o$ & $36^o$ & $54^o$ & $72^o$ & $90^o$ & $108^o$ & $126^o$ & $144^o$ & $162^o$ & $180^o$  \\
\noalign{\smallskip}\hline\noalign{\smallskip}
$0^o$&92.2&91.5&89.2&87.4&86.1&85.5&87.2&88.8&89.3&90.6&91.7 \\
$18^o$&91.3&93.5&90.9&89.5&88.2&84.6&86.2&87.1&89.8&89.4&91.8  \\
$36^o$&88.5&90.4&94.1&91.7&89.1&88.4&91.0&92.5&93.8&90.1&89.4  \\
$54^o$&87.5&88.3&89.7&91.8&90.3&87.6&90.6&91.1&90.4&89.2&87.1 \\
$72^o$&85.6&88.5&90.4&91.6&92.4&91.3&89.1&87.3&87.0&86.3&84.4 \\
$90^o$&83.0&84.7&85.3&88.6&90.8&93.7&91.7&89.6&87.4&86.2&85.1 \\
$108^o$&82.4&85.6&85.4&88.1&89.5&89.4&90.0&89.8&87.2&85.3&83.2 \\
$126^o$&84.6&86.1&88.6&90.3&89.3&88.0&89.8&91.9&87.5&85.7&83.4 \\
$144^o$&84.7&85.0&89.2&89.8&88.2&84.4&86.9&87.8&89.9&86.5&84.5 \\
$162^o$&90.5&91.2&88.8&87.2&84.6&83.4&85.1&87.4&89.7&91.5&90.2 \\
$180^o$&92.0&91.6&87.8&86.8&83.5&82.4&82.9&85.1&86.5&90.9&92.4 \\
\noalign{\smallskip}\hline
\end{tabular}
\end{table*}

We also conducted experiments under more challenging conditions such as a person
carrying an item or wearing a coat because the CASIA Dataset B includes sequences
of such conditions, which were not used in previous studies \cite{ref22}\cite{ref23}\cite{ref30}\cite{ref31}\cite{ref40}\cite{ref41}. As indicated in Tables \ref{tab:3} and \ref{tab:4}, the $CCR$ of the gender prediction was
significantly decreased under the challenging conditions of a side view or
nearside view, even when the proper classifier was applied for the specific 
viewpoint. This problem is understandable because our viewpoint-dependent 
classifiers are built upon sequences under normal walking conditions.

Moreover, for a side view or nearside view, the appearance of the person is clearly changed, both when carrying an item and when wearing a coat. The mean $\pm$ std of both $CCRs$ while carrying a bag and wearing a coat were 92.0\% $\pm$ 2.3 and 92.1\% $\pm$ 1.4, respectively. 

\begin{figure}
\centering
\includegraphics[scale=0.5]{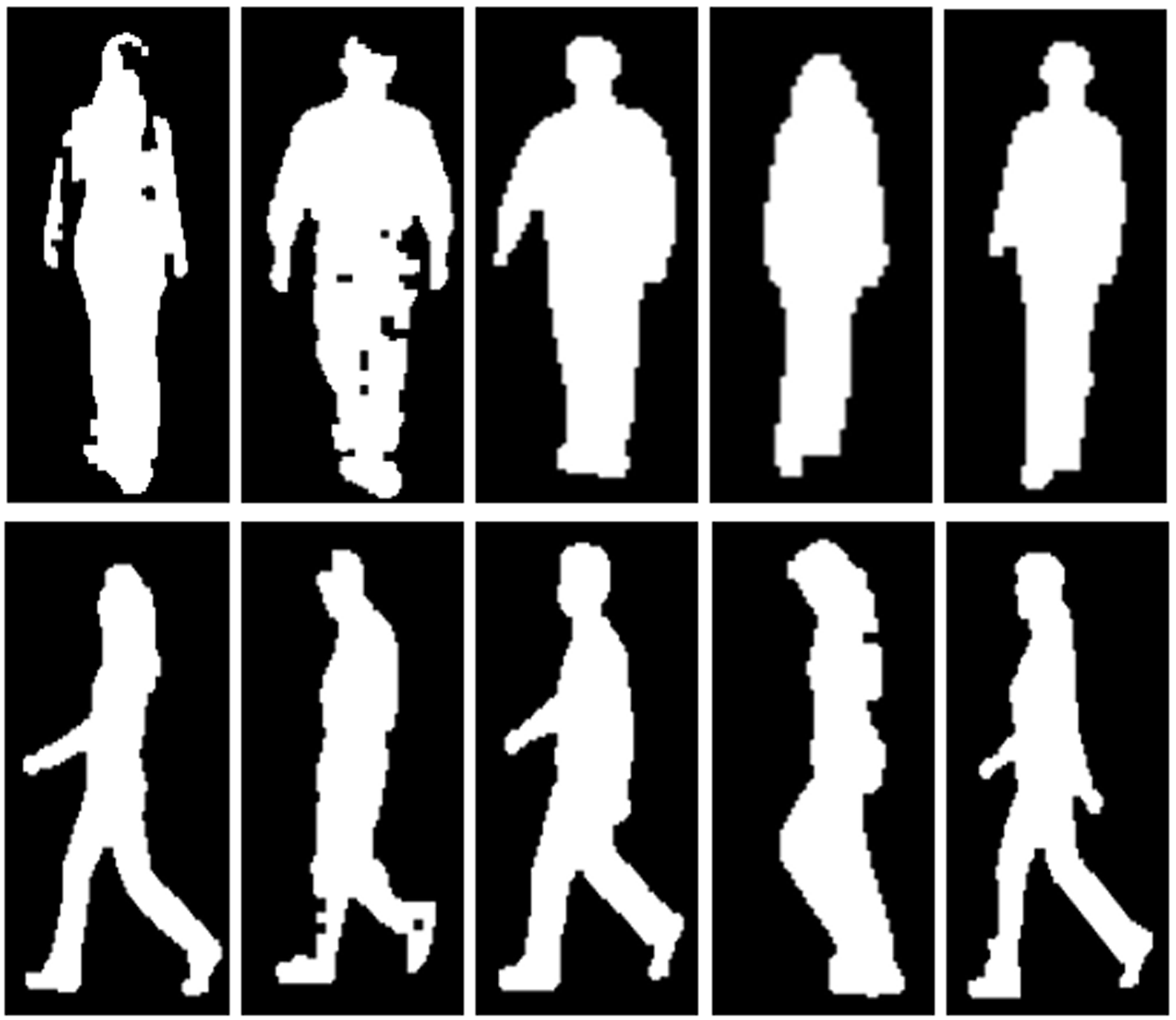}
\caption{Silhouette example of human in front view (first row) and side view (second row)}
\label{fig:7}
\end{figure}

Some examples of silhouette from the same persons in front view (top row) and side view (bottom row) are shown in Fig. \ref{fig:7}, respectively. It is interesting to notice from the figure that even in different views, the head part of the silhouette also contains the classifiable gender information (head and hair style). This is to explain that even the $90^o$ classifier is used to test the silhouette in $0^o$, the accuracy is not too low as seen in Tables 2-3-4. However, when the correct view classifier is applied, many traits for gender classification such as head and hair style, chest and back, waist and buttocks, legs  \cite{ref23} are taken into account to increase the performance.

\begin{table*}
\centering
\caption{Results of viewpoint estimation in terms of percentage for arbitrary viewpoint under normal walking conditions }
\label{tab:5}       
\begin{tabular}{llllllllllll}
\hline\noalign{\smallskip}
\multirow{2}{*}{\text{Ground-truth viewpoint}}& \multicolumn{11}{c}{Estimated viewpoint} \\
& $0^o$ & $18^o$ & $36^o$ & $54^o$ & $72^o$ & $90^o$ & $108^o$ & $126^o$ & $144^o$ & $162^o$ & $180^o$  \\
\noalign{\smallskip}\hline\noalign{\smallskip}
$0^o$&90.8&8.1&0&0&0&0&0&0&0&0&1.1 \\
$18^o$&3.4&89.5&0&0&0&0&0&0&0&8.1&0  \\
$36^o$&0&5.3&88.3&6.4&0&0&0&0&0&0&0  \\
$54^o$&0&0&4.3&82.1&11.6&0&0&0&0&0&0 \\
$72^o$&0&0&0&3.0&93.7&3.3&0&0&0&0&0 \\
$90^o$&0&0&0&0&4.4&93.2&2.4&0&0&0&0 \\
$108^o$&0&0&0&0&0&7.4&87.1&5.5&0&0&0 \\
$126^o$&0&0&0&0&0&0&1.1&88.2&10.7&0&0 \\
$144^o$&0&0&0&0&0&0&0&6.2&84.3&9.5&0 \\
$162^o$&0&0&0&0&0&0&0&0&6.4&84.5&9.1 \\
$180^o$&2.6&0&0&0&0&0&0&0&0&9.3&88.1 \\
\noalign{\smallskip}\hline
\end{tabular}
\end{table*}

\begin{table*}
\centering
\caption{CCRs (\%) of the viewpoint-dependent classifier of unknown viewing angle under different walking conditions without attachment-area removal module }
\label{tab:6}       
\begin{tabular}{llllllllllllll}
\hline\noalign{\smallskip}
\multirow{2}{*}{\text{Walking condition}}& \multicolumn{11}{c}{Classifier} \\
& $0^o$ & $18^o$ & $36^o$ & $54^o$ & $72^o$ & $90^o$ & $108^o$ & $126^o$ & $144^o$ & $162^o$ & $180^o$ & \text{Avg} & \text{Unified}  \\
\noalign{\smallskip}\hline\noalign{\smallskip}
\text{Normal walking}&98.5&99.1&98.7&97.8&99.3&99.8&98.7&98.4&98.3&98.9&99.2&98.8 &90.3\\
\text{Carrying backpack}&94.6&94.4&93.7&92.6&91.3&87.5&90.1&91.7&92.5&93.7&94.9&92.5 &85.7\\
\text{Wearing coat}&92.1&93.7&94.4&92.6&93.2&94.1&91.3&92.6&90.1&92.3&93.4&92.7 &86.1\\
\noalign{\smallskip}\hline
\end{tabular}
\end{table*}

\begin{table*}
\centering
\caption{CCRs (\%) of the viewpoint-dependent classifier of unknown viewing angle when including attachment-area removal module }
\label{tab:7}       
\begin{tabular}{lllllllllllll}
\hline\noalign{\smallskip}
\multirow{2}{*}{\text{Walking condition}}& \multicolumn{11}{c}{Classifier} \\
& $0^o$ & $18^o$ & $36^o$ & $54^o$ & $72^o$ & $90^o$ & $108^o$ & $126^o$ & $144^o$ & $162^o$ & $180^o$ & \text{Avg}  \\
\noalign{\smallskip}\hline\noalign{\smallskip}
\text{Normal walking}&98.5&99.1&98.7&97.8&99.3&99.8&98.7&98.4&98.3&98.9&99.2&98.8 \\
\text{Carrying backpack}&94.8&94.6&94.5&94.8&93.5&95.1&94.4&94.3&93.4&93.9&94.9&94.4 \\
\text{Wearing coat}&93.1&93.8&94.6&93.4&93.7&94.5&93.1&93.5&92.3&92.6&93.7&93.5 \\
\noalign{\smallskip}\hline
\end{tabular}
\end{table*}

\subsection{Viewpoint estimation test}
Viewpoint estimation is an important step in this work because it determines the DS model and classifier to be used for gender prediction. To test the accuracy of the viewpoint estimation module, we randomly selected ten sequences from a specific viewpoint under normal walking conditions from the CASIA Dataset B. This procedure was conducted as discussed in Section 3.2.2. The average percentages (for the ten sequences) of the viewpoint estimation are displayed in Table \ref{tab:5}. As can be seen, given a sequence with a specific viewpoint from the CASIA Dataset B, the viewpoint estimated from the program did not match the given viewpoint (for the given $0^o$ degree sequence, the estimated viewpoints are $0^o$, $18^o$, and $180^o$ with probabilities of 90.8, 8.1, and 1.1, respectively). This is understandable because people change their gait features while walking. Moreover, the person’s appearance from the front and rear views are similar, which results in a classification step, i.e., 90.8\% for a $0^o$ classifier, 8.1\% for an $18^o$  classifier, and 1.1\% for a $180^o$ classifier were used to obtain the gender of the individual.

After obtaining the viewing angle, the corresponding classifier was selected to conduct the gender prediction. Table \ref{tab:6} displays the $CCRs$ for an unknown viewpoint under different walking conditions, i.e., normal walking, carrying an item, and wearing a coat. The CCRs under normal walking conditions (Table \ref{tab:6}, r ow 1) are improved because the viewpoints were automatically calculated and the proper classifier was selected for the gender prediction.

For the given unknown $0^o$ viewpoint, $90.8\%$ of the image sequences are estimated at $0^o$ viewing angle, $8.1\%$ at $18^o$, and $1.1\%$ at $180^o$, respectively, as shown in Table \ref{tab:5}. Those image sequences are then sent to $0^o$, $18^o$, $180^o$ classifiers, respectively to calculate the $CCR$ at $0^o$ viewpoint. The performance using the corresponding classifiers under normal walking conditions increases the $CCRs$ from ($97.6\% \pm 0.881$) to ($98.8\% \pm 0.550$). The $CCRs$, while carrying a bag or wearing a coat (Table \ref{tab:6}, rows 2 and 3), were not significantly improved because we used classifiers trained under normal walking conditions to predict the gender.

For the aim of proving the superiority of the view-dependent design, a unified classifier of all viewpoints is trained without considering the viewing angle using the same configuration of the SVM. The experimental result is reported in the last column of Table \ref{tab:6}. As seen in the table, the average performance of view-dependent (penultimate column) increases significantly in each experimental scenario comparing to the unified classifier.

\begin{table*}
\centering
\caption{Comparison with related methods}
\label{tab:8}       
\begin{tabular}{llllll}
\hline\noalign{\smallskip}
\text{Compared methods} & \text{\#subjects} & \text{Viewpoints} & \text{Walking condition} & \text{Reported} & \text{Proposed method} \\
\noalign{\smallskip}\hline\noalign{\smallskip}

\text{Lee et al. \cite{ref42}} & \text{14 males, 10 females} & $90^o$ & \text{Normal} & $84.5\%$ & $100\%$ \\

\text{Li et al. \cite{ref44}} & \text{31 males, 31 females} & $90^o$ & \text{Normal} & $93.2\%$ & $99.8\%$\\

\text{Yu et al. \cite{ref23}} & \text{31 males, 31 females} & $90^o$ & \text{Normal} & $95.9\%$ & $99.8\%$\\

\text{Huang et al. \cite{ref43}} & \text{30 males, 30 females} & $0^o, 90^o, 180^o$  & \text{Normal} & $89.5\%$ & $99.2\%$\\

\text{Zhang De \cite{ref32}} & \text{31 males, 31 females} & $0^o, 18^o,..., 180^o$  & \text{Normal} & $98.1\%$ & $98.8\%$\\

\text{NA}& \text{31 males, 31 females} & $0^o, 18^o,..., 180^o$  & \text{Bag-carrying} & \text{NA} & $94.4\%$\\

\text{NA}& \text{31 males, 31 females} & $0^o, 18^o,..., 180^o$  & \text{Coat-wearing} & \text{NA} & $93.5\%$\\

\noalign{\smallskip}\hline
\end{tabular}
\end{table*}

\subsection{Attachment removal test}
The attachment area and noise can be removed using the procedure discussed in 
Section 3.3.2. During the testing phase, the silhouette was corrected and updated
using the attachment-area removal module. The updated version of the AGI was 
calculated based on the new version of the silhouette. As indicated in Table \ref{tab:6} and 
Table \ref{tab:7}, the $CCRs$ of a human carrying a backpack in the side view ($90^o$) is 
significantly improved with the attachment removal module (95.1\%), without the 
attachment removal module (87.5\%). More improvement in the $CCRs$ of 94.4\% $\pm$
0.564 and 93.5\% $\pm$ 0.704, compared to the cases of no attachment removal, is indicated in Table \ref{tab:7} because of the attachment-area removal module. Moreover, a significant improvement for the side view or nearside view is presented in Table \ref{tab:7}, row 2 (carrying a bag). The $CCR$ values in Table \ref{tab:7}, row 1 were not changed because in this case, the person was walking with a thin coat and not carrying any objects. The attachment-area removal module did not remove anything in this case since the distance signal was within the range of $MaDS$ and $MiDS$.
\subsection{Comparisons}
The dataset used in \cite{ref42} consists of twenty-four subjects, 14 males and 10 females, walking in normal speed and stride. The camera was placed perpendicular to their walking path. In \cite{ref23}\cite{ref44}experiments, only side-view sequences of 31 males and 31 females were collected from CASIA Dataset B for gender classification evaluation. Huang et al. in \cite{ref43} extracted only 30 males and 30 females from CASIA Dataset B in three viewing angles including $0^o$,$90^o$, and $180^o$.
Table \ref{tab:8} presents a $CCR$ comparison of the proposed method with other related works. In the side-view dataset, the proposed method attained $CCRs$ of 100\% and 99.8\% compared with 84.5\% reported in \cite{ref42} and 95.9\% reported in \cite{ref23} on a small dataset and the CASIA Dataset B under the same conditions, respectively. The proposed method was also tested on normal walking conditions in three viewing angles ($0^o,90^o$, and $180^o$) and achieved greater accuracy (99.2\% on average) compared with 89.5\% as in \cite{ref43}.

To demonstrate the effectiveness of the proposed method for gender classification with multiple-viewing angles, we conducted a test on multiple-views of the CASIA Dataset B (11 viewing angles) and obtained CCRs of 98.8\%, which is also greater than the state-of-the-art method, 98.1\% reported in \cite{ref32}.

As described in Section 4.1, this CASIA Dataset B included three categories. The first category contained videos of humans walking in a normal condition without any attachments. The two remaining categories were more challenging, containing videos of humans carrying a backpack and humans wearing a coat. Because of the attachments, the silhouettes were highly deformed, leading to significant degradation on the classification results (Table \ref{tab:6} and Table \ref{tab:7}). To the best of our knowledge, there are no experimental results reported for these two remaining datasets.

Applying the proposed module to remove the attachments, we performed experiments on these two datasets in multiple viewpoints (11 viewing angles) in the same scenario as the first category. The CCRs on the challenging dataset images indicated promising results of 94.4\% and 93.5\% for the bag-carrying and coat-wearing images, respectively, as indicated in Table \ref{tab:8}.

Further, because the proposed method uses simple operations for gender classification such as 2-dimensional signals (distance signal), and linear SVM it requires only 48 ms (20.8 frames per second) to process a frame after skipping the first 15 frames for the AGI calculation. This means that the algorithm can be applied to a surveillance application in real time.

\section{Conclusions and future works}
\begin{figure}
  \centering
  \includegraphics[scale=0.5]{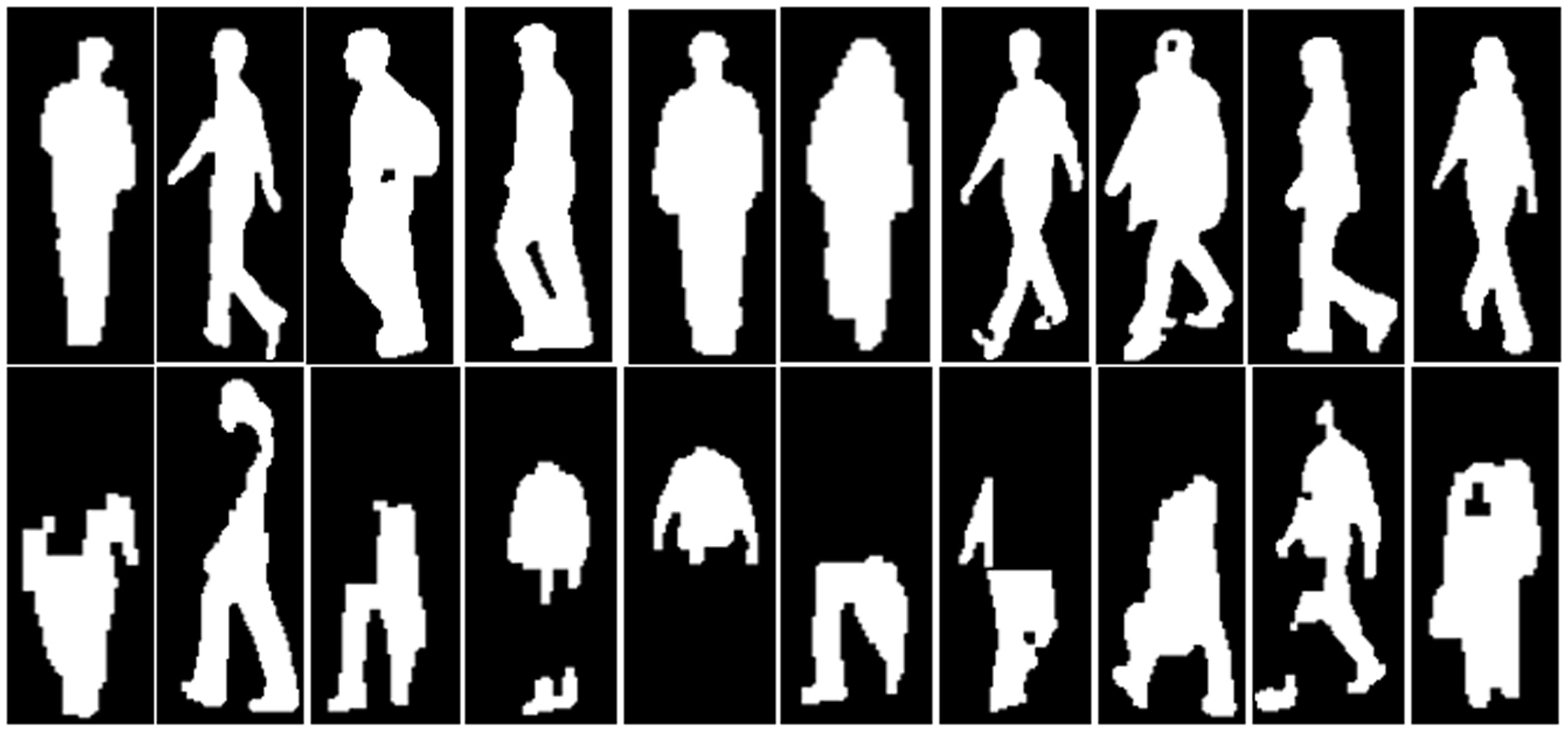}
\caption{Good and bad silhouettes obtained from the background subtraction process.}
\label{fig:8}       
\end{figure}

Gender information can be effectively obtained from a video surveillance system based on the gait feature of the subject. Instead of using a GEI, this paper employed an AGI, which is easier to calculate for a real application. To accurately predict the human gender in real applications, we created viewpoint-dependent classifiers, i.e., a VP model and a DS model. The VP model is used to estimate the viewing angle during the testing phase; any attachment area is then removed using the DS model. Finally, the gender information is provided through the use of the viewpoint-dependent classifier. A comparison with other state-of-the-art methods \cite{ref23} confirmed that the proposed method achieved a high accuracy of 98.8\% and can be applied to a real-world system. However, the results of this method depend mainly on the quality of the silhouette obtained during the background subtraction step, as shown in Fig. \ref{fig:8}. In the figure, the first row shows the samples that the method correctly classifies the gender whereas the second row shows the ones that are wrongly classified due to bad quality. As future work, we will attempt to apply the raw RGB image of a person rather than a silhouette image using deep-learning techniques because color information is an important factor for gender classification accurately predict.

\end{document}